%% file: conference_101719.tex
\documentclass[conference]{IEEEtran}
\IEEEoverridecommandlockouts
\usepackage{cite}
\usepackage{amsmath,amssymb,amsfonts}
\usepackage{algorithmic}
\usepackage{graphicx}
\usepackage{textcomp}
\usepackage{enumitem}
\usepackage{array} 
\usepackage{xcolor}
\usepackage{makecell}
\usepackage{tabularx}
\usepackage{float}
\usepackage[utf8]{inputenc}
\usepackage[T1]{fontenc}
\usepackage{newtxmath}
\def\BibTeX{{\rm B\kern-.05em{\sc i\kern-.025em b}\kern-.08em
    T\kern-.1667em\lower.7ex\hbox{E}\kern-.125emX}}
\begin{document}


\title{Arabic Dialect Classification using RNNs, Transformers, and Large Language Models: A Comparative Analysis \\

}

\author{\IEEEauthorblockN{1\textsuperscript{st} Omar A.Essameldin}
\IEEEauthorblockA{\textit{Faculty of Computer Science} \\
\textit{MSA Univeristy}\\
Giza, Egypt \\
omar.ahmed94@msa.edu.eg}
\and
\IEEEauthorblockN{2\textsuperscript{nd} Ali O.Elbeih}
\IEEEauthorblockA{\textit{Faculty of Computer Science} \\
\textit{MSA Univeristy}\\
Giza, Egypt \\
ali.osama7@msa.edu.eg}
\and
\IEEEauthorblockN{3\textsuperscript{rd} Wael H.Gomaa}
\IEEEauthorblockA{\textit{\makecell{Faculty of Computer and Artifical \\ Intelligence, Beni-Suef University}} \\
\textit{MSA Univeristy}\\
Giza, Egypt \\
wahassan@msa.edu.eg}
\and
\IEEEauthorblockN{4\textsuperscript{th} Wael F.Elsersy}
\IEEEauthorblockA{\textit{Faculty of Computer Science} \\
\textit{MSA Univeristy}\\
Giza, Egypt \\
wfarouk@msa.edu.eg}
}

\maketitle

\begin{abstract}
The Arabic language is among the most popular languages in the world with a huge variety of dialects spoken in 22 countries. In this study, we address the problem of classifying 18 Arabic dialects of the QADI dataset of Arabic tweets. RNN models, Transformer models, and large language models (LLMs) via prompt engineering are created and tested. Among these, MARBERTv2 performed best with 65\% accuracy and 64\% F1-score. Through the use of state-of-the-art preprocessing techniques and the latest NLP models, this paper identifies the most significant linguistic issues in Arabic dialect identification. The results corroborate applications like personalized chatbots that respond in users' dialects, social media monitoring, and greater accessibility for Arabic communities.

\end{abstract}

\section{Introduction}
Arabic is spoken by more than 400 million people and is a heterogeneous group of dialects subject to historical, geographical, and cultural influences. Although Modern Standard Arabic (MSA) is utilized as the formal written norm, the spoken dialects exhibit extensive divergence in phonological, morphological, syntactic, and lexical properties. Heterogeneity makes its automatic handling and computational comprehension arduous.

Dialect identification is crucial in many applications, such as machine translation, speech recognition, sentiment analysis, and social media monitoring. However, detecting dialects is very challenging due to the informal, noisy, and code-switched nature of dialectal Arabic, especially on media like Twitter.

This work centers on classifying 18 dialects with the QADI dataset, which contains 440,000 Arabic tweets. Somalia, Djibouti, Comoros, and Mauritania dialects are excluded because they have poor representation, but the dataset includes the rest of the Arab nations with rich regional diversity.
We propose a machine learning pipeline integrating preprocessing, feature engineering, and deep learning models—specifically, RNNs, Transformers, and LLMs. With analysis of linguistic patterns within social media text, this study enriches dialect identification and real-world applications such as personalized chatbots, dialect-aware NLP systems, and better accessibility for Arabic-speaking populations.

\section{Related Work}
    Arabic dialect identification has been extensively studied with different models, datasets, and methodological approaches. Besides the key models and datasets introduced in Table~\ref{tab:related_work_summary}, several comparative benchmarks and support materials have been instrumental to this area. Research tends to compare its outcomes against baseline models like MARBERT \cite{abdul-mageed2021marbert}, mBERT \cite{devlin2019bert}, RoBERTa \cite{liu2019roberta}, and AraBERT \cite{antoun2020arabert}, emphasizing the challenge of variation between dialects. Datasets such as PADIC \cite{luo-etal-2016-automatic} and AOC \cite{zaidan2014arabic} have significantly enhanced the landscape such that more sophisticated assessment of dialectical variation can be achieved. Traditional feature engineering methods such as Word2Vec and GloVe embeddings, and simple models such as Support Vector Machines (SVMs) and TF-IDF representations have played a crucial role in the initial stages of research. Though the supporting models and data are not necessarily reflected in the table below, they present background data that is crucial in grasping how Arabic dialect detection systems evolved. A summary of the key contributions being explored within the respective literature is given in Table~\ref{tab:related_work_summary}. The table details the best performing models, suggested approaches, evaluation metrics, and data sets utilized within the studies reviewed, offering a consolidated point of reference for ongoing study.

    \begin{table*}[ht]
        \centering
        \small
        \caption{Related Work Comparative Table}
        \vspace{4pt}
        \renewcommand{\arraystretch}{1.7}
        \begin{tabularx}{\textwidth}{|>{\centering\arraybackslash}X|
                                         >{\centering\arraybackslash}X|
                                         >{\centering\arraybackslash}X|
                                         >{\centering\arraybackslash}X|
                                         >{\centering\arraybackslash}X|}
            \hline
            \textbf{Paper Name} & \textbf{Best Model} & \textbf{Proposed Model} & \textbf{Metric} & \textbf{Datasets} \\
            \hline
            \makecell{Arabic Dialect \\ Identification \\ in the Wild \\ \cite{abdelali2020arabic}} & AraBERT \cite{antoun2020arabert} & \makecell{AraBERT, \\ Mazajak, \\ mBERT} & \makecell{\textbf{60.6\%} \\ F1-score} & QADI \cite{abdelali2020arabic} \\
            \hline
            \makecell{A Three-Stage \\ Neural Model \\ for Arabic \\ Dialect Identification \\ \cite{mohammed2023three}} & MDA-BERT & \makecell{Three-Stage \\ Neural Model \\ (Word \\ Weighting)} & \makecell{\textbf{67.16\%} \\ Accuracy} & \makecell{QADI \cite{abdelali2020arabic} \\ MADAR \cite{bouamor2018madar} \\ NADI \cite{bouamor2020nadi}} \\
            \hline 
            \makecell{Enhancing Arabic \\ Dialect Detection \\ on Social Media: \\ A Hybrid Model with \\ Attention \\ \cite{yafooz2024enhancing}} & \makecell{Hybrid Model \\ (LSTM, BiLSTM, \\ Logistic \\ Regression + \\ Attention)} & \makecell{Hybrid Model \\ (LSTM, BiLSTM, \\ Logistic Regression, \\ Attention Layer)} & \makecell{\textbf{88.73\%} \\ Accuracy \\ (Twitter)} & \makecell{Twitter \\ MADAR \cite{bouamor2018madar} \\ NADI \cite{bouamor2020nadi} \\ QADI \cite{abdelali2020arabic}} \\
            \hline
            \makecell{Machine Learning \\ Based Approach \\ for Arabic Dialect \\ Identification \\ \cite{nayel-etal-2021-machine}} & \makecell{Complement \\ Naïve Bayes \\ (CNB)} & \makecell{TF-IDF \\ + CNB \\ Model} & \makecell{\textbf{18.72\%} \\ Macro F1 \\ (Country)} & NADI \cite{bouamor2020nadi} \\
            \hline
            \makecell{Systematic \\ Literature Review \\ of Dialectal Arabic: \\ Identification \\ and Detection \\ \cite{elnagar2021systematic}} & (Survey) & No Proposed Model & - & \makecell{MADAR \cite{bouamor2018madar} \\ PADIC \cite{luo-etal-2016-automatic} \\ NADI \cite{bouamor2020nadi} \\ AOC \cite{zaidan2014arabic}} \\
            \hline
            \makecell{Advancing Arabic \\ Dialect Detection \\ with Hybrid  Stacked \\ Transformer Models \\ \cite{saleh2025advancing}} & \makecell{Stacking \\ AraBERTv02 \\ + Dialectal\\ -XLM -R + RF} & \makecell{Stacked \\ Transformer \\ Ensemble \\ (BERT + \\ XLM-R + RF)} & \makecell{\textbf{93.18\%} \\ F1-score \\ (IADD)} & \makecell{Shami \\ IADD} \\
            \hline
        \end{tabularx}
        \normalsize
        \label{tab:related_work_summary}
    \end{table*}

\section{Dataset}
    \subsection{QADI}
        One of the Arabic dialect detection challenges of consequence is that there do not exist large, well-balanced datasets covering all 22 Arab-speaking countries. Most existing datasets are small-grained, genre-limited, or size-limited and consist of greatly imbalanced data or include excessive Modern Standard Arabic (MSA).
        To overcome this difficulty, we utilized the QADI dataset \cite{abdelali2020arabic}, a large corpus created in 2021 using the Twitter Streaming API. The corpus is made up of 440,000 tweets in 18 Arabic dialects, thus ranking among the largest datasets to date. Mauritania, Somalia, Djibouti, and Comoros dialects are not included because of data unavailability. QADI still presents a representative and varied sample that can be used for dialect classification despite this restriction.


\section{Methodology}
\begin{figure*}[h]
        \centering
        \includegraphics[width=0.9\linewidth]{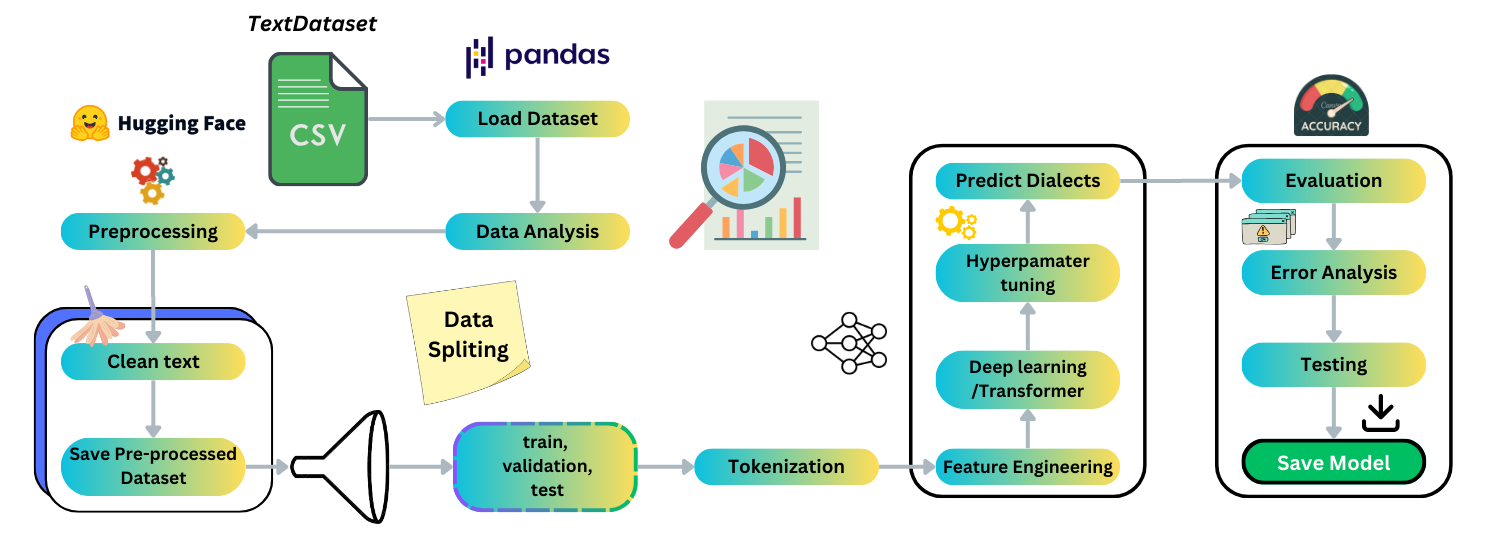}
        \caption{Proposed Model}
        \label{fig:model_architecture}
    \end{figure*}

    We propose a hybrid RNN and Transformer-based Arabic dialect classification pipeline, illustrated in Figure~\ref{fig:model_architecture}. The pipeline consists of preprocessing, dataset splitting, tokenization, model training, and evaluation.

    RNN-based models (LSTM, GRU) are built on top of AraVec 3.0 word embeddings, followed by recurrent layers, an attention mechanism, and a softmax classifier. Every word $w$ in vocabulary $\mathcal{V}$ is embedded into a vector $v_w$ as:
    
    \begin{equation}
    v_w = f(w), \quad f: \mathcal{V} \rightarrow \mathbb{R}^d
    \end{equation}
    
    where $f$ is the embedding function and $d = 100$ is the embedding dimension. Afterwards, self-attention mechanism is applied to dynamically assess the significance of each token, which is expressed as:
    
    \begin{equation}
    \text{Attention}(Q, K, V) = \text{softmax}\left(\frac{QK^\top}{\sqrt{d_k}}\right)V
    \end{equation}
    
    Here, $Q$, $K$, and $V$ are query, key, and value matrices derived from input embeddings, and $d_k$ is the dimension of the key. This formulation also appears in Transformer models to allow each token to attend to the remaining tokens in the sequence.
    
    Transformer models (AraBERT, CAMelBERT-DA, E5, and MARBERT) bypass manual feature engineering by fine-tuning with token IDs and attention masks, keeping model-internal representations and contextual dependencies intact. This is favored over inputting precomputed embeddings in that it allows end-to-end training of all layers. The modular design enables uniform assessment of various models. The following sections describe each component of the pipeline, beginning with preprocessing.

    \subsection{Preprocessing}
        \subsubsection {cleaning text}
            In this stage, we used a sequence of text sanitization methods inspired by the preprocessing approach introduced by Abdelali et al. (2020) \cite{abdelali2020arabic}, utilizing their open-source script. The key steps included substituting URLs and user IDs with generic placeholders, stripping diacritics to eliminate redundancy, and normalizing characters by unifying various representations of the same letter. Special Unicode characters were eliminated, redundant emojis and redundant characters were fixed, and word boundaries in multiple languages were fixed to enhance tokenization. Unnecessary whitespace was also eliminated. The preprocessed dataset was subsequently divided into training, validation, and test sets without changing the original dialect balance. This preprocessing facilitated the creation of a more uniform dataset, eliminating extraneous variables and computation needs while retaining vital linguistic aspects necessary for detecting dialects.
    \subsection{Deep learning models}
        \subsubsection{Tokenization} 
            Tokenization divides text into individual tokens or units. For our Arabic dialect classification framework, we utilized the Treebank tokenizer, which is known to deal with Arabic's morphologically rich nature. It tokenizes text on whitespace and punctuation but preserves cases like contractions and multi-word terms.
        \subsubsection{Feature Engineering}
            Before model training, text was transformed into numerical representations via word embeddings. We utilized AraVec 3.0 \cite{soliman2017aravec}, a pre-trained Arabic embedding model based on the Word2Vec CBOW algorithm. Each word was mapped to a 100-dimensional vector reflecting semantic relationships, providing dense input features that improved model understanding of linguistic patterns.
    \subsection{RNN-Based models}
        Both GRU and LSTM models utilized RNN-based architectures to learn temporal and linguistic relationships in Arabic text for dialect classification. The models were trained on the entire dataset using Google Colab Pro. AraVec 3.0 embeddings were used to vectorize input text. Both models also included self-attention mechanisms from Keras to increase attention to significant features within the sequences. Additionally, dropout layers and early stopping with model checkpointing were implemented in an attempt to negate overfitting and maintain optimal model weights. Final classification utilized a dense layer with softmax activation, in accordance with the total number of target dialect categories.
        \vspace{4pt}

    \subsubsection{\textbf{LSTM and GRU}}
        
        Within the framework of Arabic dialect identification, two types of recurrent neural network (RNN) architectures—Long Short-Term Memory (LSTM) and Gated Recurrent Unit (GRU)—were utilized for modeling temporal dependencies and capturing complex morphological patterns. The LSTM architecture was implemented with two stacked layers: the first layer fed the entire sequence to a self-attention mechanism, while the next layer gave the final hidden state. This setup was followed by a dropout layer and a dense softmax classifier. The architectural model of the GRU model was identical to that of its counterpart, substituting LSTM cells with GRU cells, which yield better computational efficiency thanks to fewer gating mechanisms. Both models used AraVec 3.0 word embeddings for input representation and used early stopping with model checkpointing in order to keep track of validation loss. Learning rate and dropout rate were both altered in an attempt to optimize performance within resource constraints. The LSTM model could achieve 52\% accuracy on the overall classification task, while the GRU model matched this performance with faster training and less memory use, making it more suitable for low-resource environments.
         
        \vspace{5pt}
        
    \subsection{Transformers models}
        \vspace{2pt}
        To classify Arabic dialects, we trained four transformer models—AraBERT Base Twitter v02 \cite{antoun2020arabert}, E5-Multilingual-Large \cite{wang2024multilingual}, MARBERTv2 \cite{abdul-mageed-etal-2021-arbert}, and CAMeLBERT-DA \cite{inoue-etal-2021-interplay}—on Google Colab Pro with the Hugging Face Transformers library. We trained all models to classify 18 Arabic dialects with dedicated training, validation, and test splits. A learning rate of 2e-5 was applied across all models, although the learning rate for MARBERTv2 was subsequently adjusted to 1e-5 following the emergence of early overfitting signs observed after the second epoch. The training configuration involved the AdamW optimizer alongside a weight decay of 0.01, with evaluation performed at each epoch and logging every 500 steps. A custom evaluation function was used when calculating the accuracy and F1-score for each epoch. Each model was equipped with a certain tokenizer to create input IDs and attention masks, casting the datasets into (input IDs, attention mask, labels) format for supervised fine-tuning. Other settings coalesced around using a two-stage gradient accumulation scheme, 0.1 warmup ratio, and judicious application of mixed precision (fp16 or bf16) depending on model size and GPU availability. The best-performing model based on validation loss was restored after training concluded.

        All the models were trained for multi-class classification with the categorical cross-entropy loss function, a metric that computes the divergence between the actual label distribution and the predicted probability distribution. It is mathematically formulated as:
        \begin{equation}
        \mathcal{L}_{CE} = -\sum_{i=1}^{N} y_i \log(\hat{y}_i)
        \end{equation}

        Here, $y_i$ is the actual label, while $\\hat{y}_i$ is the predicted probability for class $i$.
        
        Both models were fed token IDs and attention masks and evaluated with accuracy and macro F1-score. Logging was done at each 500 steps.
        
        \vspace{5pt}
    \subsubsection{\textbf{AraBERT base v02 twitter}}
         We pre-trained AraBERT Base Twitter v02 \cite{antoun2020arabert}, which is a 110M parameter model pre-trained exclusively on Arabic tweets, on an L4 GPU with 22.5GB VRAM. We pre-trained the model for 15 epochs, batch size 128, for both training and validation. Preprocessing was done by utilizing AraBERT's dedicated preprocessor module tailored for social media text normalization. Mixed precision (fp16) was not used as model size was manageable. Training lasted approximately 6 hours, with overfitting occurring after the 4th epoch, however the best model was saved at the end. AdamW optimization was used without quantization, and validation was performed every epoch to monitor performance.
        \vspace{5pt}
    \subsubsection{\textbf{E5-Multilingual-Large}}
       The E5-Multilingual-Large model \cite{wang2024multilingual} with 1.24 billion parameters was fine-tuned on L4 GPU to accommodate the larger model size and memory needs. As a multilingual retrieval and classification model, E5 was inherently able to handle cross-dialectal variation. To achieve training efficiency, we have enabled bnb 8-bit quantization and mixed-precision (fp16), reducing memory usage significantly and speeding up computation. Training was completed for 5 epochs with a training batch size of 64 and validation batch size of 128, with gradient accumulation to effectively have a batch size of 128. Max gradient norm was set to 1.0 to prevent gradient explosion. Although larger, E5 completed training in about 2.5 hours with performance comparable to the smaller models.
       \vspace{5pt}
    \subsubsection{\textbf{MARBERTv2}} 
        For the MarBERTv2 model \cite{abdul-mageed-etal-2021-arbert}, fine-tuning on a 90/5/5 train/val/test split was performed. Training was performed with a learning rate of 1.5e-5, train batch size and eval batch size of 64 and 128 respectively, with (bf16) precision and a linear scheduler for 5 epochs. A DataCollatorWithPadding was utilized in conjunction with the model's tokenizer to accommodate variable input length. Early stopping was triggered in the 4th epoch due to overfitting as eval loss started stabilizing. The Trainer was configured to track validation loss and utilize the best-performing model, with gradient accumulation of 2, warmup ratio of 0.1, and weight decay of 0.01 to enhance generalization.
        \vspace{5pt}
    \subsubsection{\textbf{CAMeLBERT-DA}} 
        CAMelBERT-DA \cite{inoue-etal-2021-interplay}, a 110M parameter dialect-adapted BERT model specifically trained on Arabic dialectal data, was fine-tuned using a T4 GPU with high RAM. The model was trained for 5 epochs with a batch size of 64 for training and 128 for validation. Given its specialization for dialectal Arabic, CAMelBERT-DA showed strong performance despite being lighter than E5. Early stopping with a patience of 2 epochs was applied to mitigate overfitting, and no mixed-precision (fp16) or quantization was used. Training was completed in approximately 1.5 hours. Despite using a lower-tier GPU, CAMelBERT-DA achieved performance comparable to AraBERT, demonstrating its efficiency in Arabic dialect classification tasks.
        \vspace{5pt}
        
    \subsection{LLMs}
        To address Arabic dialect detection, we employed large language models (LLMs) and refined our methodology through staged prompt engineering. The process included three stages: zero-shot learning, zero-shot with prompt engineering techniques, and few-shot learning. In zero-shot learning, prompts contained only queries, enhanced by techniques like Tree of Thoughts and Chain of Words. Few-shot learning introduced a few labeled examples to improve model understanding. We used three models via free-tier APIs, selecting a sample of 9,900 records to fit request limits. To optimize, a unified zero-shot prompt was used with 10 sentences per request, splitting the dataset into three batches of 3,300 records to prevent API failures. For few-shot learning, prompts were shortened to 5 sentences to stay within the models token-per-minute cap, with 2 to 4 second delays between requests to avoid limits. Each model was evaluated across four phases: basic zero-shot, enhanced zero-shot with prompt engineering (COW and ToT), few-shot learning.
    \vspace{8pt}
    \subsubsection{\textbf{Gemma-3-27b-it}}
    Gemma\cite{gemma_2025} was obtained via the Google GenAI API. In zero-shot, it recorded 28\% accuracy and 28\% F1-score. Subsequently, after prompt engineering techniques such as Chain of Words and Tree of Thoughts were applied, the accuracy stayed the same while the F1-score decreased marginally to 27\%. In few-shot learning, its performance deteriorated even more, with accuracy and F1-score reducing to 25
    \vspace{8pt}
    \subsubsection{\textbf{Gemini-2.0-Flash-Lite}}
    Gemini-2.0-Flash-Lite, also accessible via the Google GenAI API, scored 31\% accuracy and 31\% F1-score under the zero-shot setting. When fine-tuned with prompt engineering methods like Chain of Words and Tree of Thoughts, the model had the same level of performance scores. But in the few-shot setting, accuracy and F1-score both fell to 28\%, indicating a marginal dip in performance.
    \vspace{8pt}
    \subsubsection{\textbf{QWEN-2.5-32b}}
    \cite{qwen2.5} was experimented with on the Groq Cloud API. It started off with 19\% accuracy and 18.7\% F1-score in the zero-shot run. After applying prompt engineering techniques, it managed to achieve slightly better accuracy at 20\% and F1-score of 19.5\%. As in the case of Gemma and Gemini, QWEN also showed a marginal but positive gain, but overall performance remained poor.

         
\section{Results}

    \subsection{Recurrent Models Performance}
    The GRU and LSTM models were trained over the entire QADI dataset using AraVec 3.0 embeddings (100-dimensional vector). Both produced 52\% accuracy, with LSTM performing marginally better in F1-score (51\% vs 50.5\%), as demonstrated in Table~\ref{tab:model_comparsion}. GRU provided improved computational performance, leaning towards resource-limited environments like Google Colab.
    
    \subsection{Transformer-Based Models}
    AraBERT-V02-Twitter, E5-Multilingual-Large, MARBERTv2, and CAMeLBERT-DA were trained on the dataset using Hugging Face's Trainer API. MarBERT performed best (65\% accuracy, 64\% F1-score) with AraBERT close behind at 63\%, and both E5 and CAMeLBERT achieved 60\%. Confusion matrix of MAREBRTV2 (Figure~\ref{fig:marbert}, revealed dialectal overlap problems despite good overall performance.
    
    \subsection{Dialect-Level Insights}
        As indicated in Table~\ref{tab:marbert_report}, MARBERTv2 performed well on high-resource dialects Egyptian (EG), Moroccan (MA), and Lebanese (LB) with F1-scores of 0.85, 0.76, and 0.74 respectively. These dialects had the advantage of larger sample sizes and more unique lexical features that allowed the model to learn more distinct representations.
        
        However, the model was confronted with difficulty in handling dialects like Yemeni (YE), Jordanian (JO), and Omani (OM), which recorded lower F1-scores of 0.39, 0.46, and 0.52 respectively. The drop in performance can be explained by high lexical similarities to neighboring dialects, which caused them to be harder to discriminate against.
        
\begin{figure}[h]
        \centering
        \includegraphics[width=1.1\linewidth]{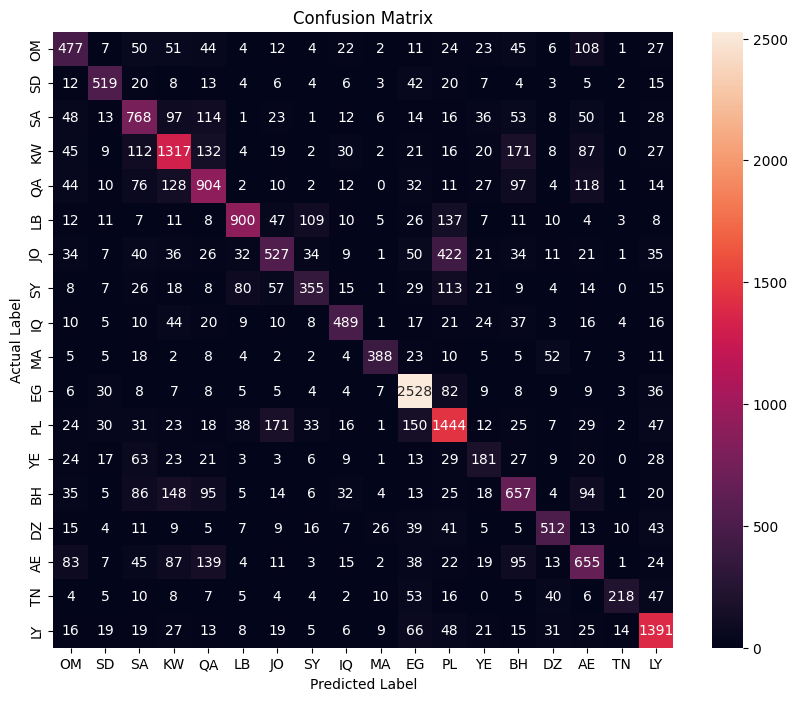}
        \caption{Confusion matrix of finetuned MARBERTv2}
        \label{fig:marbert}
    \end{figure}

        The confusion matrix in Figure~\ref{fig:marbert} confirms these findings. For example, Yemeni (YE) and Jordanian (JO) labels have heavy confusion with Palestinian (PL) and Saudi (SA) predictions, respectively. Likewise, Bahraini (BH) instances are frequently confused with Kuwaiti (KW) and Saudi (SA). In contrast, Egyptian (EG) and Levantine dialects such as PL and LB exhibit robust diagonal dominance, reflecting correct predictions. These confusion patterns reveal that the dialects with near geographic or socio-linguistic relations overlap more in representation space, causing issues for the model's classification boundaries.
    
    \subsection{Large Language Models and Prompt Engineering}
    Gemma-3-27b-it, Gemini-2.0-Flash-Lite, and Qwen2.5-32b were evaluated using zero-shot and few-shot prompt engineering. Gemini did best in zero-shot (31\% accuracy). Few-shot learning hindered the performance of all the models because Arabic-specific tokenizer issues ruined input structure, which thwarted the purpose of carefully crafted prompts, and because the dataset had numerous classes, just one example from a single dialect was included in the prompt, so the prompt was lengthy.
    \begin{table}[h]
        \centering
        \renewcommand{\arraystretch}{1.5} 
        \setlength{\tabcolsep}{7pt} 
        \caption{Results}
        \vspace{4pt}
         \begin{tabular}{|c|c|c|c|}
            \hline
                \textbf{Model} & \textbf{Accuracy (\%)} & \textbf{F1 score (\%)} & \textbf{Dataset}\\
            \hline
                LSTM & 52 & 51 & QADI \\
            \hline
                GRU & 52 & 50.5 & QADI\\
            \hline 
                AraBERT & 63 & 62 & QADI\\
            \hline
                \textbf{MARBERTv2} & \textbf{65} & \textbf{64} & QADI \\
            \hline
                E5-Multilingual-Large & 60 & 60 & QADI \\
            \hline
                CAMelBERT & 60 & 60 & QADI \\
            \hline
         \end{tabular}
         \label{tab:model_comparsion}
    \end{table}
    \begin{table}[h]
    \centering
    \renewcommand{\arraystretch}{1.4}
    \caption{MarBERTv2 classification report}
    \vspace{4pt}
    \setlength{\tabcolsep}{4pt}
    \begin{tabular}{|c|c|c|c|c|}
        \hline
        \textbf{Dialect} & \textbf{precision (\%)} & \textbf{recall (\%)} & \textbf{f1-score} & \textbf{support} \\
        \hline
            OM & 0.5288 & 0.5196 & 0.5242 & 918 \\
            \hline
            SD & 0.7310 & 0.7489 & 0.7398 & 693 \\
            \hline
            SA & 0.5486 & 0.5958 & 0.5712 & 1289 \\
            \hline
            KW & 0.6443 & 0.6513 & 0.6478 & 2022 \\
            \hline
            QA & 0.5711 & 0.6059 & 0.5880 & 1492 \\
            \hline
            LB & 0.8072 & 0.6787 & 0.7374 & 1326 \\
            \hline
            JO & 0.5553 & 0.3930 & 0.4603 & 1341 \\
            \hline
            SY & 0.5936 & 0.4551 & 0.5152 & 780 \\
            \hline
            IQ & 0.6986 & 0.6573 & 0.6773 & 744 \\
            \hline
            MA & 0.8273 & 0.7004 & 0.7586 & 554 \\
            \hline
            EG & 0.7987 & 0.9133 & 0.8522 & 2768 \\
            \hline
            PL & 0.5783 & 0.6873 & 0.6281 & 2101 \\
            \hline
            YE & 0.3969 & 0.3795 & 0.3880 & 477 \\
            \hline
            BH & 0.5042 & 0.5206 & 0.5123 & 1262 \\
            \hline
            DZ & 0.6975 & 0.6589 & 0.6777 & 777 \\
            \hline
            AE & 0.5113 & 0.5186 & 0.5149 & 1263 \\
            \hline
            TN & 0.8226 & 0.4910 & 0.6150 & 444 \\
            \hline
            LY & 0.7593 & 0.7939 & 0.7762 & 1752 \\
        \hline
        \textbf{accuracy} & \_ & \_ & \textbf{0.6467} & 22003 \\
        \hline
        \textbf{macro avg} & 0.6430 & 0.6094 & \textbf{0.6213} & 22003 \\
        \hline
        \textbf{weighted avg} & 0.6474 & 0.6467 & \textbf{0.6435} & 22003 \\
        \hline
    \end{tabular}
    \label{tab:marbert_report}
\end{table}

\section{Conclusion \& Future work}
    This paper presented a comprehensive research on Arabic dialect classification with RNN-based models, Transformer models, and LLMs with prompt engineering. LSTM and GRU performed moderately, and LSTM performed slightly better in F1-score.
    Among the Transformers, MARBERTv2 outperformed other models with 65\% accuracy and 64\% F1-score, which validated the strength of domain-specific pretraining. Dialect-level analysis showed excellent performance in high-resource dialects but performed poorly for low-resource dialects.
    LLMs proved to be somewhat acceptable in zero-shot dialect classification and then further deteriorated in few-shot learning due to poor Arabic tokenization, which warped example structures. Despite advanced prompt engineering, tokenizer, and context size constraints remained a primary bottleneck.
    Preprocessing techniques — normalization, diacritics removal, and mention/URL processing — were crucial to boost model performance over noisy Twitter text.

    Future work needs to try to expand data sets to include low-resource dialects and less-represented regions, as well as incorporating phonological and syntactic features to allow for finer-grained classification. Investigation into model ensembling and multi-branch architectures will hopefully increase dialectal robustness. Equally important is the development of tokenizers and pre-training objectives that are optimized for Arabic in order to maximize large language model (LLM) performance. Additional work on cross-lingual transfer learning and domain-adaptive learning can continue to enhance the development of more generalizable and inclusive dialect identification systems.

\bibliographystyle{IEEEtran}
\input{conference_101719.bbl}

\end{document}